%% file: ijcai20.tex
\documentclass{article}
\usepackage{ijcai20}

\usepackage{times}

\usepackage{soul}
\usepackage{url}
\usepackage[utf8]{inputenc}
\usepackage[small]{caption}
\usepackage{graphicx}
\usepackage{amsmath}
\usepackage{booktabs}
\usepackage{latexsym}
\usepackage{epsfig}
\usepackage{amssymb}
\usepackage{mathtools}
\usepackage{bm}
\usepackage{xcolor}
\usepackage{multirow}
\usepackage{caption}
\usepackage{array}
\usepackage{dsfont}
\usepackage[hidelinks]{hyperref}

\DeclareMathOperator*{\softmax}{softmax}
\DeclareMathOperator*{\fc}{fc}

\title{Learning to Discretely Compose Reasoning Module Networks\\for Video Captioning}

\author{
Ganchao Tan$^{1*}$\and
Daqing Liu$^1$\footnote{Equal Contribution}\and
Meng Wang$^2$\And
Zheng-Jun Zha$^{1}$\footnote{Corresponding Author}\\
\affiliations
$^1$University of Science and Technology of China\\
$^2$Hefei University of Technology\\
\emails
\{tgc1997, liudq\}@mail.ustc.edu.cn,
eric.mengwang@gmail.com,
zhazj@ustc.edu.cn
}

\begin{document}

\maketitle

\input{0_abstract.tex}
\input{1_introduction.tex}
\input{2_related_work.tex}
\input{3_approach.tex}
\input{4_experiments.tex}
\input{5_conclusion.tex}

{\small
\bibliographystyle{ijcai20}
\bibliography{citations}
}

\end{document}

%% file: 0_abstract.tex
\begin{abstract}
Generating natural language descriptions for videos, \textit{i.e.}, video captioning, essentially requires step-by-step reasoning along the generation process.
For example, to generate the sentence ``a man is shooting a basketball'', we need to first locate and describe the subject ``man'', next reason out the man is ``shooting'', then describe the object ``basketball'' of shooting.
However, existing visual reasoning methods designed for visual question answering are not appropriate to video captioning, for it requires more complex visual reasoning on videos over both space and time, and dynamic module composition along the generation process.
In this paper, we propose a novel visual reasoning approach for video captioning, named Reasoning Module Networks (RMN), to equip the existing encoder-decoder framework with the above reasoning capacity.
Specifically, our RMN employs 1)~three sophisticated spatio-temporal reasoning modules, and 2)~a dynamic and discrete module selector trained by a linguistic loss with a Gumbel approximation.
Extensive experiments on MSVD and MSR-VTT datasets demonstrate the proposed RMN outperforms the state-of-the-art methods while providing an explicit and explainable generation process. Our code is available at \url{https://github.com/tgc1997/RMN}.
\end{abstract}

%% file: 1_introduction.tex
\section{Introduction}
Video captioning, the task aims to automatically generate natural language descriptions for videos, has received increasing attention in computer vision and machine learning. Even though our community achieves the significant advance in visual recognition~\cite{he2016deep,ren2015faster} and natural language understanding~\cite{bahdanau2014neural}, video captioning is still a very challenging task and far away from satisfactory for it not only requires a thorough understanding of the input videos, but also requires step-by-step visual reasoning along the generation process.

\input{figures/fig1.tex}

As illustrated in Figure~\ref{fig:intro}, to describe the video, we human may have the following reasoning process: 1) identifying the subject to describe of the video, \textit{i.e.}, ``man'', 2) inferring what is the man doing, \textit{i.e.}, ``shooting'', 3) identifying what is the object of ``shooting'', \textit{i.e.}, ``basketball'', 4) generating the final description ``a man is shooting a basketball'' by inserting several function words. In a nutshell, there are three fundamental reasoning mechanisms involved: 1) Locate one region to generate visual words, 2) Relate pairwise regions to generate action words, 3) Generate function words according to the language context.

Most existing video captioning methods~\cite{venugopalan2015sequence,donahue2015long} follow the encoder-decoder framework, where a CNN is employed as an encoder to produce the video features and an RNN is employed as a decoder to generate the captions. Those methods usually neglect the nature of the above human-level reasoning, thus hurting the explainability of the generation process.
Even though there are some recent works have explored the visual reasoning in visual question answering~\cite{andreas2016neural,hu2017learning,9009518} and visual grounding~\cite{cirik2018using,liu2019learning,hong2019learning} by decomposing the questions or referring expressions into a linear or tree reasoning structure with several neural modules, the situation in the video captioning is more challenging because 1) unlike still images, videos contain richer visual content thus requiring more complex visual reasoning over both space and time, 2) unlike questions or referring expressions which are given in advance, the video descriptions are not available during the inference. Therefore, the model must dynamically compose the reasoning structure along the generation process.

To tackle the above two challenges, we propose a novel video captioning framework named Reasoning Module Networks (RMN).
Firstly, to perform visual reasoning over both space and time, our RMN employs three fundamental spatio-temporal reasoning modules: a)~\textsc{Locate} module to locate one single region over the video by a spatial-temporal attention, thus generating the visual words, \textit{e.g.}, ``man'' and ``basketball'' in Figure~\ref{fig:intro}; 
b)~\textsc{Relate} module to relate pairwise regions over the video by first detecting the object of each frame, and then modeling the action by pairing two frames, thus generating the action words, \textit{e.g.}, ``shooting'' in Figure~\ref{fig:intro}; and
c)~\textsc{Func} module to generate the function words according to the language context, \textit{e.g.}, ``a'' and ``is'' in Figure~\ref{fig:intro}.
Secondly, to compose the reasoning structure along the generation process, our RMN employs a dynamic and discrete module selector.
To jointly train both the three modules and the selector in an end-to-end manner, we adopt the recently proposed Gumbel approximation~\cite{jang2016categorical} to make the discrete sampling process differentiable, and then constraint the module selector with a linguistic loss of part-of-speech (POS) tag labels.

We validate the effectiveness of the proposed RMN by conducting extensive experiments on two widely-used datasets MSVD~\cite{chen2011collecting} and MSR-VTT~\cite{xu2016msr}. RMN outperforms the state-of-the-art video captioning methods on most metrics.
Qualitative results indicate the generation process is explicit and explainable.

Our main contributions are three-fold:
1) We propose a novel framework named reasoning module networks (RMN) for video captioning with three \textit{spatio-temporal} visual reasoning modules;
2) We adopt a \textit{discrete} module selector to \textit{dynamically} compose the reasoning process with modules;
3) Our RMN achieves new state-of-the-art performance with an explicit and explainable generation process.

%% file: figures/fig1.tex
\begin{figure}[t]
	\centering
	\includegraphics[width=\linewidth]{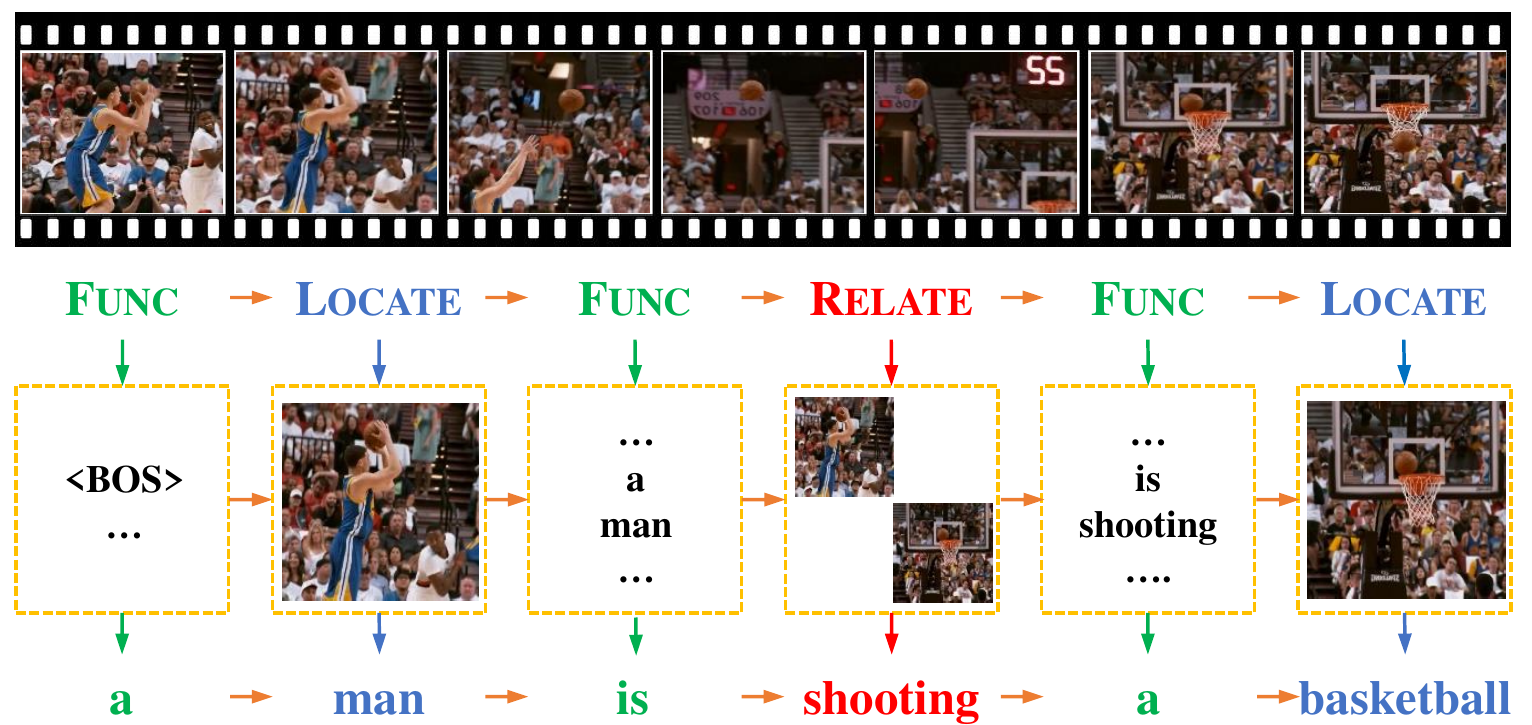}
	\caption{The caption generation process of the proposed Reasoning Module Networks (RMN). At each step, RMN first makes a dynamic and discrete selection from three fundamental reasoning modules, \textit{i.e.}, \textsc{Locate}, \textsc{Relate}, and \textsc{Func}, and then executes the corresponding reasoning module to generate the word. Specifically, \textsc{Locate} module locates one region to generate the visual words, \textsc{Relate} module relates pairwise regions over both space and time to generate action words, \textsc{Func} module generates those function words according to the language context.}
	\label{fig:intro}
\end{figure}

%% file: 2_related_work.tex
\section{Related Work}
\subsection{Video Captioning}
There are two main directions to solve the video captioning problem. In the early stage, template-based methods~\cite{kojima2002natural,guadarrama2013youtube2text}, which first define a sentence template with grammar rules and then aligned subject, verb and object of the sentence template with video content, were widely studied. Those methods are hard to generate flexible language due to the fixed syntactic structure of the predefined template.
Benefit from the rapid development of deep neural networks, the sequence learning methods~\cite{venugopalan2015sequence,yao2015describing,pan2017video} are widely used to describe the video with flexible natural language, most of these methods are based on the encoder-decoder framework.
\cite{venugopalan2015sequence} proposed S2VT model which regards the video captioning task as a machine translation task. ~\cite{yao2015describing} introduced a temporal attention mechanism to assign weights to the features of each frame and then fused them based on the attention weights. \cite{li2017mam,chen2019motion} further applied spatial attention mechanisms on each frame.

Recently, \cite{wang2019controllable} and \cite{hou2019joint} proposed to leverage Part-of-Speech (POS) tags to boost video captioning. \cite{wang2019controllable} encodes the predicted POS sequences into hidden features, which further guides the generation process. \cite{hou2019joint} mixes word probabilities of multiple components at each timestep conditioned on the inferred POS tags. However, both of them lack the reasoning capability for rich video content. On the contrary, we propose three well-designed reasoning module networks that correspond to three fundamental reasoning mechanisms.

\subsection{Neural Module Networks}
Neural module networks is a general framework that explicitly models the compositionality of languages by decomposing the network into neural modules. It has been widely used in visual question answering and visual grounding.
\cite{andreas2016neural} employs an off-the-shelf parser to parse the questions into a tree, leading to brittleness caused by parsing errors.
\cite{hu2017learning} trains an RNN to decode the language into the sequence, thus requiring extra human annotations.
\cite{hu2018explainable} removes language parser and additional annotations by using a soft and continuous module layout.
However, the case of video captioning is more complex since there are no fully observed captions during inference thus all the above methods are not applicable. To this end, we design a dynamic module selector to construct the reasoning procedure step-by-step during the generation process.

Recently, the pioneering works~\cite{liu2018context,8684270,yang2019learning,tian2019image} are trying to adopt neural module networks into image captioning.
However, their modules are designed to produce several types of features and their module compositions relied on a soft attention mechanism.
As a contrast, our RMN employs several sophisticated spatio-temporal reasoning modules to perform more complex visual reasoning over videos, and designs a discrete and dynamic module selector to make a selection at each step, making the generation process explicit and explainable.

%% file: 3_approach.tex
\section{Approach}

\input{figures/fig2.tex}
In this section, we will describe the proposed Reasoning Module Networks (RMN) in more detail.
Figure~\ref{fig:networks} gives a walk-through example of our RMN at timestep $t$. Our RMN can be divided into four stages: Encoding (Section~\ref{sec:3.1}), Module Reasoning (Section~\ref{sec:3.2}), Module Selection (Section~\ref{sec:3.3}), and Decoding (Section~\ref{sec:3.4}). In Section~\ref{sec:3.5}, we detail the joint training strategy for both reasoning modules and the module selector.

\subsection{Encoder}
\label{sec:3.1}
For the given video of $N$ frames, we first represent it with three types of features, \textit{i.e.}, appearance features $\bm{V_a}$ extracted from a 2D-CNN, object features $\bm{V_o}$ extracted from a R-CNN on each frame, and motion features $\bm{V_m}$ extracted from a 3D-CNN. Note that to model temporal information for those visual features, we have post-processed $\bm{V_a}$ and $\bm{V_m}$ with Bi-LSTMs. And $\bm{V_o}$ contains an additional dimension on space.

For the previous generated words $\{w_1, \cdots, w_{t-1}\}$, we encode them by a LSTM (denoted as enLSTM) which takes the global visual feature $\bar{\bm{v}}$, the last generated word embedding vector $\bm{e}_{t-1}$, and the last timestep hidden state $\bm{h}^{de}_{t-1}$ of deLSTM (cf. Eqn.~\eqref{eq:delstm}) as input:
\begin{equation}
    \bm{h}^{en}_t = \textbf{enLSTM}(\bar{\bm{v}}, \bm{e}_{t-1}, \bm{h}^{de}_{t-1}).
\label{eq:enlstm}
\end{equation}
It is worth noticing that since $\bm{h}^{en}_t$ encodes rich information from history, it will be used to guide the reasoning process and the module selection.

\subsection{Reasoning Modules}
\label{sec:3.2}
As mentioned in the Introduction, there are three fundamental reasoning mechanisms involved in video captioning.
Therefore, we design three corresponding reasoning modules. In the following, we will first introduce the attention functions used in the modules, and then describe each module in detail.

\subsubsection{Attention Functions}
We follow the widely used additive attention formulation~\cite{bahdanau2014neural}:
\begin{equation}
    \textbf{A}(\bm{V}, \bm{q}) = \softmax(\bm{w}^T_1 \tanh (\bm{W}_2 \bm{V} + \bm{W}_3 \mathds{1}^T \bm{q})) \bm{V},
\end{equation}
where $\bm{V}$ and $\bm{q}$ are values and queries of attention, $\bm{w}_1$, $\bm{W}_2$, and $\bm{W}_3$ are trainable parameters, $\mathds{1}$ is an all-one vector.
Note that, the attention can be executed on any dimensions, therefore, we further define $\textbf{AoS}(\cdot)$ as the Attention over the dimension on Space, and define $\textbf{AoT}(\cdot)$ as the Attention over the dimension on Time.

\subsubsection{\textsc{Locate} Module}
It is designed to generate visual words, \textit{e.g.}, ``man'' and ``basketball''. Generating this type of words requires the model to attend on one specific region over both space and time. Therefore, we first apply an Attention over Space (AoS) for object features $\bm{V_o}$ and then apply an Attention over Time (AoT) together with $\bm{V_a}$, formally:
\begin{equation}
    \bm{v^l_t} = \textbf{AoT}(\textbf{AoS}(\bm{V_o}, \bm{h}^{en}_t) \oplus \bm{V_a}, \bm{h}^{en}_t),
\end{equation}
where $\oplus$ denotes concatenate operation.

\subsubsection{\textsc{Relate} Module}
It is designed to generate action words, \textit{e.g.}, ``riding'' and ``shooting''. Modeling action requires reasoning over both space and time.
Take Figure~\ref{fig:intro} as an example, to generate the word ``shooting'', we must be aware that the man is holding a basketball, and then notice that the basketball is flying to the basket, finally we can inference that the man is ``shooting''.
It means we must connect two scenes at different time.
To achieve this reasoning mechanism, we insert a pairwise interaction (\textit{i.e.}, concatenate every possible pair of two tensors) between the AoS and AoT, formally:
\begin{equation}
\begin{split}
    \bm{v^r_t} & = \textbf{AoT}(\textbf{P}(\bm{M}, \bm{M}), \bm{h}^{en}_t), \\
    \bm{M} & = \textbf{AoS}(\bm{V_o}, \bm{h}^{en}_t) \oplus \bm{V_m},
\end{split}
\end{equation}
where $\textbf{P}(\cdot, \cdot)$ denotes the pairwise interaction function that $\textbf{P}_{ij}(\bm{A}, \bm{B})=\bm{A}_i \oplus \bm{B}_j$.

\subsubsection{\textsc{Func} Module}
It is designed to generate function words to complete the whole sentences. Since the function words only require language information, thus we propose to recall the history decoder cell states of deLSTM (cf. Eqn.~\eqref{eq:delstm}) to generate the current words by an AoT, formally:
\begin{equation}
\begin{split}
    \bm{v^f_t} & = \textbf{AoT}(\bm{C}, \bm{h}^{en}_t), \\
    \bm{C} & = [\bm{c}^{de}_1, \cdots, \bm{c}^{de}_{t-1}].
\end{split}
\label{eq:func}
\end{equation}

So far, we have detailed the three proposed reasoning modules. Next, we will describe how to make a selection of those modules at each timestep.

\subsection{Module Selector}
\label{sec:3.3}
As illustrated in Figure~\ref{fig:networks}, the module selection consists of two steps. First, we calculate a score for each module to measure the probability of each module could be selected.
Second, we sampling one determined module based on those scores with Gumbel Softmax~\cite{jang2016categorical} strategy.

In detail, we formulate the scoring function $\textbf{S}(\cdot, \cdot)$ as:
\begin{equation}
    \textbf{S}(\bm{h}, \bm{v}) = \fc(\tanh(\fc(\bm{h}) + \fc(\bm{v}))).
\end{equation}
Thus we can get $s^l_t=\textbf{S}(\bm{h}^{en}_t, \bm{v}^l_t)$, $s^r_t=\textbf{S}(\bm{h}^{en}_t, \bm{v}^r_t)$, and $s^f_t=\textbf{S}(\bm{h}^{en}_t, \bm{v}^f_t)$ for the three \textsc{Locate}, \textsc{Relate}, and \textsc{Func} modules, respectively.

After that, the straightforward selection method is directly choosing the module with the maximum score.
However, it will block the gradients off the module selector for the function $\arg\max$ is non-differentiable.
To this end, we apply a relaxation between forward pass and backward pass, formally:
\begin{equation}
\begin{split}
    & \mathrm{Forward}: \bm{z}_t = \arg\max(\log([s^l_t, s^r_t, s^f_t]) + G), \\
    & \mathrm{Backward}: \tilde{\bm{z}}_t = \softmax((\log([s^l_t, s^r_t, s^f_t]) + G)/\tau),
\end{split}
\end{equation}
where we take an approximation that relaxes the discrete one-hot decision vector $\bm{z}_t$ into a continuous $\tilde{\bm{z}}_t$.
$G$ is the Gumbel noise drawn from i.i.d. Gumbel(0, 1). $\tau$ is a temperature parameter to control the strength of softmax.
Please refer to \cite{jang2016categorical} for more mathematical details.

With this decision vector $\bm{z}_t$, we can get the final visual reasoning result as:
\begin{equation}
    \bm{v}_t = \bm{z}_t \otimes [\bm{v}^l_t, \bm{v}^r_t, \bm{v}^f_t],
\label{eq:vt}
\end{equation}
where $\otimes$ denotes inner product. It is worth noticing that $\bm{z}_t$ is an one-hot vector, thus only one determined module will be selected in the forward pass.

\subsection{Decoder}
\label{sec:3.4}
For each timestep $t$, we decode the final visual reasoning result by a LSTM (denotes as deLSTM):
\begin{equation}
    \bm{h}_t^{de}, \bm{c}_t^{de} = \textbf{deLSTM}(\bm{v}_t, \bm{h}_t^{en}).
\label{eq:delstm}
\end{equation}
Recall that the hidden and cell states were used in Eqn.~\eqref{eq:enlstm} and Eqn.~\eqref{eq:func}.
Further, we decode the word probability $\bm{p}_t$ as:
\begin{equation}
    \bm{p}_t = \softmax(\textbf{MLP}(\bm{v}_t \oplus \bm{h}_t^{en} \oplus \bm{h}_t^{de})),
\end{equation}
where $\textbf{MLP}$ is two-layers with $\tanh$ as activation function.

\subsection{End-to-End Training}
\label{sec:3.5}
Thanks to the Gumbel-Softmax strategy, our RMN model can be trained in an end-to-end manner.
Given a video and the corresponding ground-truth caption $\{w^*_t\}$ for $t \in [1, T]$, where $T$ is the sentence length, the objective is to minimize the cross-entropy loss function given by:
\begin{equation}
    \mathcal{L}_{cap} = - \sum _{t=1}^{T}\log\bm{P}_t(w^*_t).
\end{equation}

Further, to ensure the module selector sticking on the linguistic structure, we propose to supervise the module selection by extra Part-of-Speech (POS) tag labels. Accordingly, we apply a Kullback-Leibler Divergence (KLD) loss to make the predicted module decision vector $\tilde{\bm{z}}$ and the ground-truth tag distribution $S^*=\{s^*_t\},\, t\in[1, T]$ as close as possible, formally:
\begin{equation}
    \mathcal{L}_{pos} = - \sum_{t=1}^{T} \mathrm{KLD}(\tilde{\bm{z}}_t \| \mathtt{one\_hot}(s^*_t)).
\end{equation}

To collect the ground-truth $S^*$, we first detected POS tag labels by Spacy Tagging Tool\footnote{https://spacy.io}, and then assigned [NN*] and [JJ*] to the \textsc{Locate} module, [VB*] to the \textsc{Relate} module, and the other to the \textsc{Func} module.

Therefore, the overall loss function of our RMN model is given by:
\begin{equation}
    \mathcal{L} = \mathcal{L}_{cap} + \lambda \mathcal{L}_{pos},
\end{equation}
where $\lambda$ is a trade-off weight.

%% file: figures/fig2.tex
\begin{figure*}[t]
	\centering
	\includegraphics[width=\linewidth]{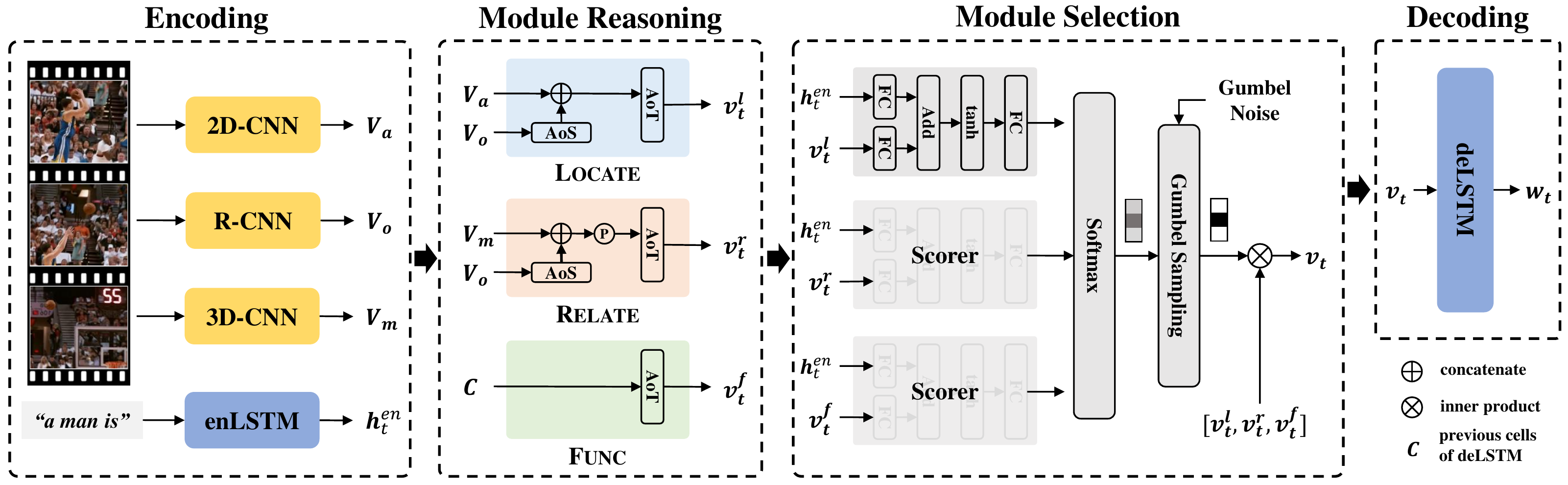}
	\caption{The overview of our proposed Reasoning Module Networks (RMN) which consists of four stages.
	At the encoding stage (cf. Section~\ref{sec:3.1}), we represent the given video by several visual features and encode previously generated words into the hidden state.
	At the module reasoning stage (cf. Section~\ref{sec:3.2}), we perform three fundamental visual reasoning over both space and time by \textsc{Locate}, \textsc{Relate}, and \textsc{Func} modules.
	At the module selection stage (cf. Section~\ref{sec:3.3}), we dynamically and discretely select one determined reasoning module to produce the final reasoning result.
	At the decoding stage (cf. Section~\ref{sec:3.4}), we decode the reasoning result into word.}
	\label{fig:networks}
\end{figure*}

%% file: 4_experiments.tex
\section{Experiments}

\input{tables/ablation.tex}

\subsection{Datasets and Metrics}
We conduct experiments on two widely used video captioning datasets with several standard evaluation metrics to verify the effectiveness of our proposed method.

\subsubsection{Datasets}
\noindent{\textbf{MSVD}.} The MSVD dataset \cite{chen2011collecting} consists of 1,970 short video clips selected from Youtube, where each one depicts a single activity in the open domain, and each video clip is annotated with multi-lingual captions. Since we only consider the English captions in this paper, each video clip has roughly 41 descriptions. To be consistent with previous works, we split the dataset to 3 subsets, 1,200 clips for training, 100 clips for validation, and the remaining 670 clips for testing.

\noindent{\textbf{MSR-VTT}.} The MSR-VTT \cite{xu2016msr} is a large-scale dataset for the open domain video captioning, which consists of 10,000 video clips from 20 categories, and each video clip is annotated with 20 English sentences by Amazon Mechanical Turks. There are about 29,000 unique words in all captions. Following the existing works, we use the standard splits, namely 6,513 clips for training, 497 clips for validation, and 2,990 clips for testing.

\subsubsection{Evaluation Metrics}
We use several widely used automatic evaluation metrics to evaluate the quality of the generated captions, \textit{i.e.}, BLEU \cite{papineni2002bleu}, METEOR \cite{banerjee2005meteor}, CIDEr \cite{vedantam2015cider}, ROUGE-L \cite{lin-2004-rouge}. Most of these metrics are originally proposed for machine translation or image captioning, the higher score indicates better quality of the captions.

\subsection{Implementation Details}
\subsubsection{Dataset Preprocessing} We first convert all captions to lower case and remove punctuations, then we truncate the captions with more than 26 words and zero pad the captions with less than 26 words. The vocabulary size is set to 7,351 for MSVD and 9,732 for MSR-VTT with removing the words appear less than twice and five times respectively.

\subsubsection{Feature Extraction}
In our experiments, we use InceptionResNetV2 (IRV2) \cite{szegedy2017inception} as 2D CNN and I3D \cite{carreira2017quo} as 3D CNN to extract appearance features and motion features respectively, then we equally-spaced 26 features for each video. The IRV2 is trained on ILSVRC-2012-CLS image classification dataset \cite{Russakovsky2015ImageNetLS} and the I3D is trained on Kinetics action classification dataset \cite{1705.06950}. We adopt Faster-RCNN \cite{ren2015faster} which is trained by \cite{Anderson2017up-down} to extract 36 region features for each frame (26 equally-spaced frames for each video).

\subsubsection{Training Details}
Our model is optimized by Adam Optimizer ~\cite{kingma2014adam}, the initial learning rate is set to 1e-4. For the MSVD dataset, the hidden size of the LSTM is 512 and the learning rate is divided by 10 every 10 epochs. For the MSR-VTT dataset, the hidden size of the LSTM is 1,300 and the learning rate is divided by 3 every 5 epochs. During testing, we use beam search with size 2 for the final caption generation.

\subsection{Ablation Study}

\input{figures/pie.tex}
\input{figures/word_cloud.tex}
\input{figures/example.tex}

As shown in Table~\ref{table:ablation}, we compare our RMN against a set of other ablated models with various settings:
(1) \textbf{RMN (\textsc{Locate})}: the model that only deployed with Locate module;
(2) \textbf{RMN (\textsc{Relate})}: the model that only deployed with Related module;
(3) \textbf{RMN (S)}: the model that softly fuses three modules, \textit{i.e.}, $\bm{z}_t$ in Eqn.~\eqref{eq:vt} is computed as $\softmax(\log([s^l_t, s^r_t, s^f_t])$;
(4) \textbf{RMN (H)}: the model that hard selects one of three modules;
(5) \textbf{RMN (S+L)}: the model that softly fuses three modules and trained by linguistic loss; 
(6) \textbf{RMN (H+L)}: the model that hard selects one of three modules and trained by linguistic loss.
According to the results shown in Table~\ref{table:ablation}, we have the following observations.

\subsubsection{Effect of Reasoning Modules}
RMN (\textsc{Relate}) consistently outperforms RMN (\textsc{Locate}) on both MSVD and MSR-VTT. This is because the \textsc{Relate} module models much richer visual action information than \textsc{Locate}, leading to performance improvement over RMN (\textsc{Locate}).

RMN (S) outperforms RMN (\textsc{Locate}) and RMN (\textsc{Relate}) under most metrics. This indicates that by fusing different types of reasoning mechanisms, the model can generate better video descriptions.

\subsubsection{Effect of Gumbel Strategy}
Comparing RMN (H) with RMN (\textsc{Locate}) and RMN (\textsc{Relate}), we observe opposite results on MSVD and MSR-VTT.
To figure out the reasons for this observation, we count the proportion of the three modules occupied in Figure~\ref{fig:pie}. We can find that without the linguistic loss, RMN (H) only employs two modules, \textit{i.e.}, \textsc{Locate} module and \textsc{Relate} module, indicating the failure of training.

Therefore, the reason for the opposite results is that RMN (H) failed in training on MSVD while succeeded on MSR-VTT.
Actually, even though Gumbel strategy can relax the discrete decision to continuous decision, it still requires large training examples.
Therefore, RMN (H) underperforms RMN (\textsc{Locate}) and RMN (\textsc{Relate}) on MSVD which only contains 1,200 training samples, while outperforming them on MSR-VTT which contains 6,513 training samples.

\input{tables/sota.tex}

Similarly, even though RMN(S) surpasses RMN(H) under some metrics, but if we employ the linguistic loss to make sure both soft fusion and hard module selector works well, RMN (H+L) outperforms RMN (S+L) with a margin.

\subsubsection{Effect of Linguistic Loss}
Comparing RMN (H+L) with RMN (H), as well as comparing RMN (S+L) with RMN (S), we find that the linguistic loss consistently improves the performance, indicating the importance of linguistic information for video captioning.

We can also observe the consistent results in Figure~\ref{fig:pie}. By imposing the linguistic loss, we can enforce the module selector sticking with the linguistic structure, and the proportion of each module is approaching to the ground-truth.

Further, we carried out the word cloud statistics as shown in Figure~\ref{fig:wordcloud}.
We can find that the three modules have a strong pattern: the \textsc{Locate} module mainly generates visual words, \textit{e.g.}, ``man'' and ``person''; the \textsc{Relate} module mainly generates action words, \textit{e.g.}, ``playing'' and ``talking''; the \textsc{Func} module mainly generates function words, \textit{e.g.}, ``about'' and ``to''. It indicates that the model has learned linguistic knowledge correctly.

\subsection{Comparison with State-of-the-Art}

We compare our proposed RMN with the most recent state-of-the-art methods on MSVD and MSR-VTT datasets. According to whether they leverage the POS labels, we categorize them into two groups: 1) traditional encoder-decoder based models MAM-RNN~\cite{li2017mam}, RecNet~\cite{wang2018reconstruction}, MARN~\cite{pei2019memory}, and OA-BTG~\cite{zhang2019object}, and 2) POS strengthened model POS-CG~\cite{wang2019controllable} and Mixture~\cite{hou2019joint}.

As shown in Table~\ref{table:sota}, we can find that: 1) the methods that leverage POS labels outperform the methods without POS information, and 2) our proposed reasoning module network outperforms the methods with POS labels on most metrics and achieves new state-of-the-art.
It is worth noting that CIDEr is proposed for captioning task specifically, and is considered more consistent with human judgment. Our model achieves the best CIDEr score on both datasets, which demonstrates our RMN model can generate captions that more in line with human descriptions.

\subsection{Qualitative Analysis}
In this section, we would like to investigate the generation process of our model by qualitative results.
Here we provide some video captioning examples in Figure~\ref{fig:example}.
As expected, our module selection is explicit and reasonable, for example, we compose ``man'', ''flute``, and ``phone'' with the \textsc{Locate} module, ``playing'', ``riding'', and ``talking'' with \textsc{Relate} module, and ``a'', ``of'', and ``the'' with \textsc{Func} module.

The above observations suggest that the generation process of the proposed RMN is totally explicit, since at each timestep the module selects one determined module to produce the current word. In addition, each module is well-designed to perform spatio-temporal visual reasoning, \textit{e.g.}, the spatio-temporal attention for \textsc{Locate} module and pairwise interaction reasoning over both space and time for \textsc{Relate} module, indicating that our model is explainable.

%% file: tables/ablation.tex
\begin{table*}[t]
\centering
\small
\begin{tabular}{@{}|c|c|c|c|c|c|c|c|c|c|c|c|c|c|@{}}
\hline
\multicolumn{6}{|c|}{Settings} & \multicolumn{4}{c|}{MSVD} & \multicolumn{4}{c|}{MSR-VTT} \\ \hline
Model &\textsc{Locate} & \textsc{Relate} & \textsc{Func} & Discrete & $\mathcal{L}_{pos}$ & B@4 & R & M & C & B@4 & R & M & C \\ \hline
RMN (\textsc{Locate})&$\checkmark$ & & & & & 52.5 & 73.1 & 35.8 & 90.0 & 40.7 & 60.5 & 27.7 & 46.9 \\ 
RMN (\textsc{Relate})& &$\checkmark$ & & & & 52.8 & 73.1 & 36.1 & 90.0 & 40.1 & 60.6 & 28.1 & 47.1 \\ 
RMN (S)&$\checkmark$ & $\checkmark$ & $\checkmark$ & & & 53.2 & 73.2 & 35.8 & 90.8 & 41.0 & 60.6 & 28.0 & 47.2 \\ 
RMN (H)&$\checkmark$ & $\checkmark$ & $\checkmark$ & $\checkmark$ & & 51.5 & 72.0 & 35.1 & 88.4 & 41.9 & 60.9 & 28.1 & 48.1 \\ 
RMN (S+L)&$\checkmark$ & $\checkmark$ & $\checkmark$ & & $\checkmark$ & 52.5 & 72.7 & 36.1 & 92.8 & 42.1 & 61.3 & 28.3 & 49.1 \\ 
RMN (H+L)&$\checkmark$ & $\checkmark$ & $\checkmark$ & $\checkmark$ & $\checkmark$ & \textbf{54.6} & \textbf{73.4} & \textbf{36.5} & \textbf{94.4} & \textbf{42.5} & \textbf{61.6} & \textbf{28.4} & \textbf{49.6} \\ \hline
\end{tabular}
\caption{The performance of ablated models with various settings on MSVD and MSR-VTT datasets. B@4, R, M, C denote BLEU-4, ROUGE\_L, METEOR, CIDEr, respectively.}
\label{table:ablation}
\end{table*}

%% file: figures/pie.tex
\begin{figure}[t]
	\centering
	\includegraphics[width=\linewidth]{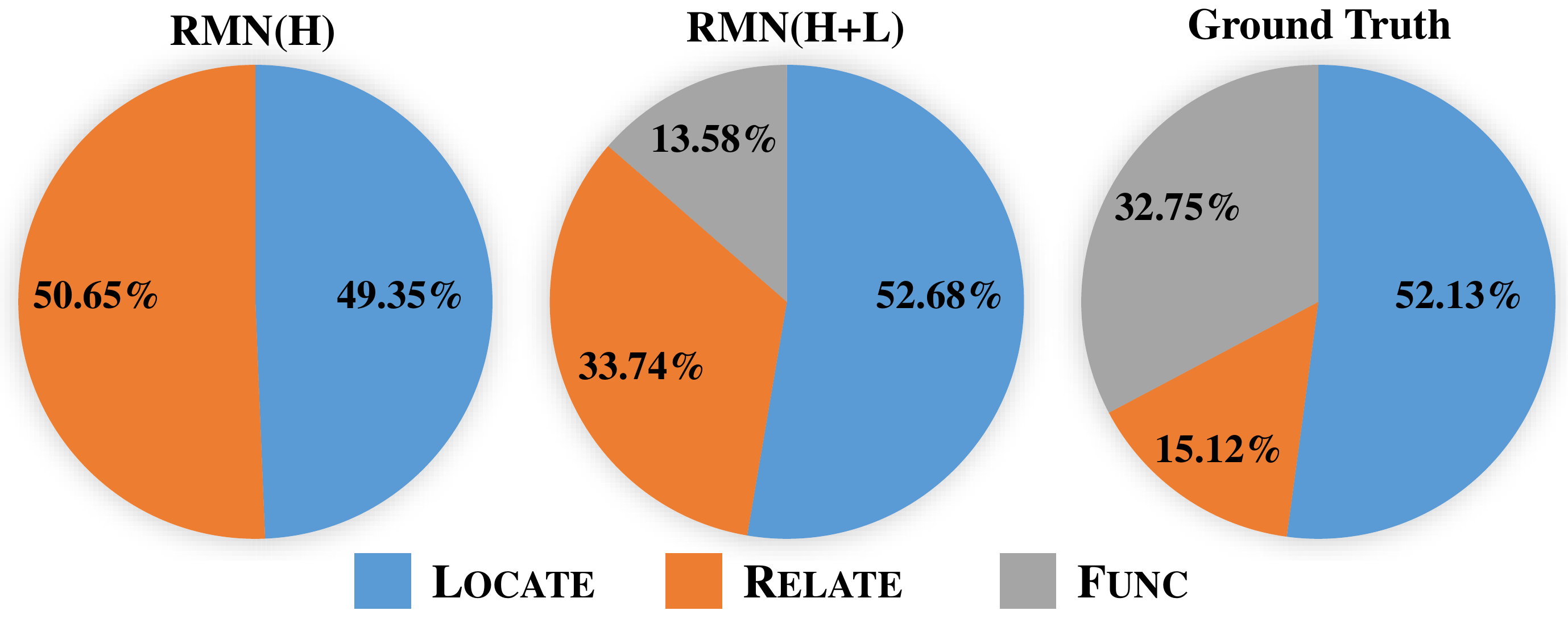}
	\caption{The proportion of each module occupied in RMN (H), RMN (H+L), and the Groud-Truth on test set of MSVD.}
	\label{fig:pie}
\end{figure}

%% file: figures/word_cloud.tex
\begin{figure}[t]
	\centering
	\includegraphics[width=\linewidth]{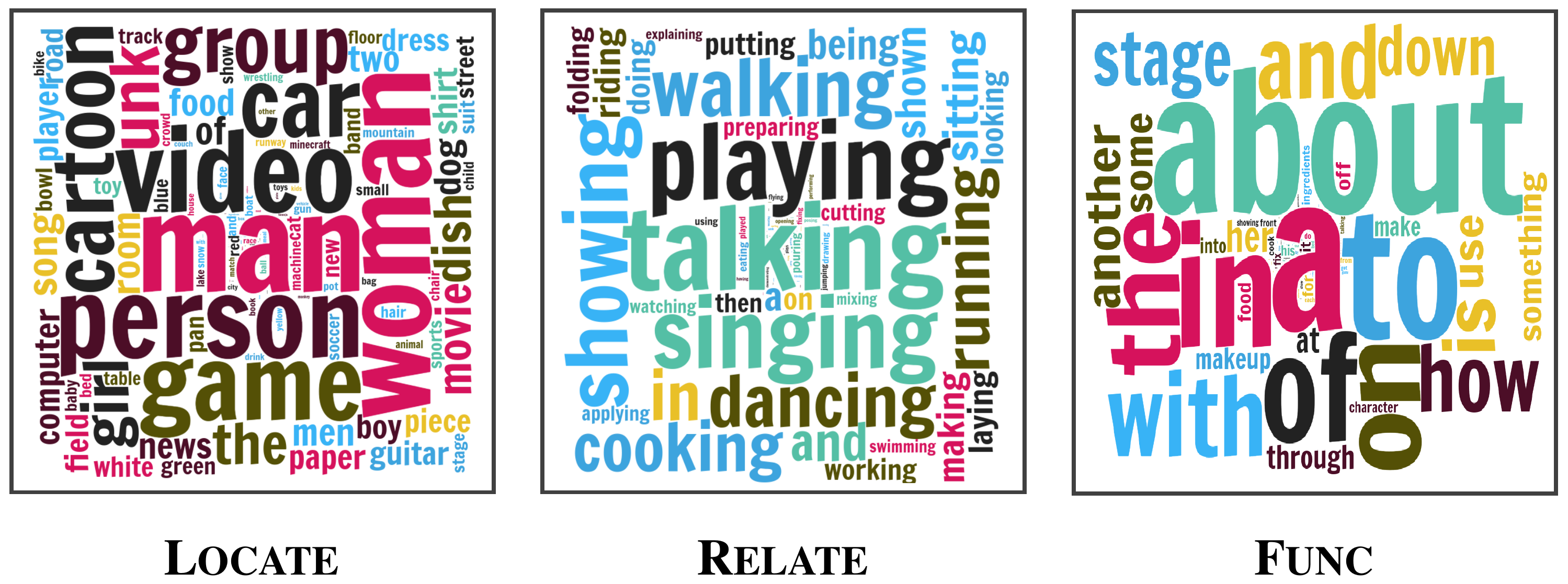}
	\caption{Word cloud visualizations of words generated by each module. We can find that the \textsc{Locate} module mainly generate visual words, the \textsc{Relate} module mainly generate action words, and the \textsc{Func} module is likely to generate function words. Experiments conducted on MSVD of RMN (H+L).}
	\label{fig:wordcloud}
\end{figure}

%% file: figures/example.tex
\definecolor{locate}{RGB}{0,112,192}
\definecolor{relate}{RGB}{255,39,39}
\definecolor{func}{RGB}{0,176,80}

\begin{figure*}[t]
	\centering
	\includegraphics[width=\linewidth]{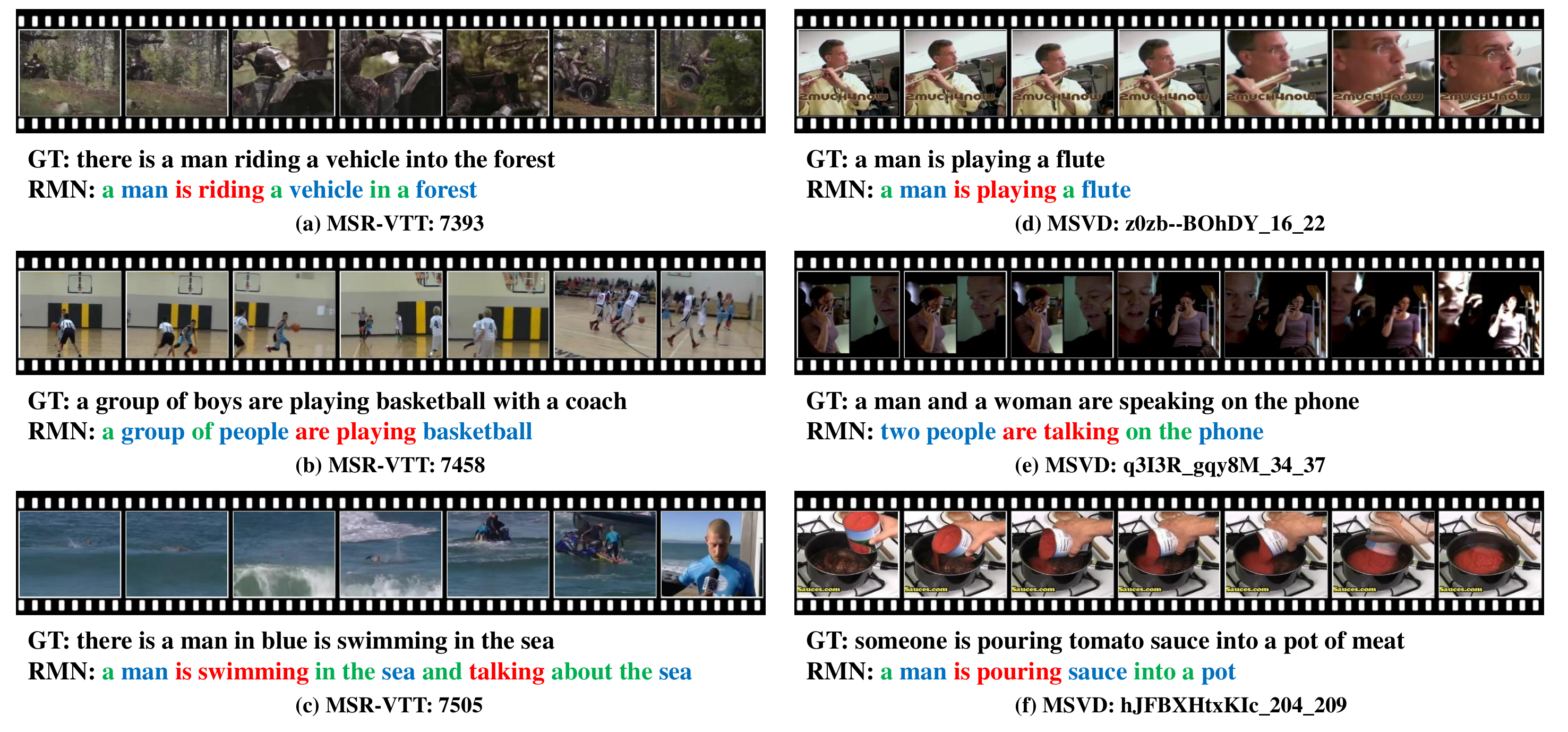}
	\caption{Visualization of some video captioning examples on MSVD and MSR-VTT (better view in color). The first line in each example is one of the ground truth captions and the second line is generated by our RMN. Word in blue, red, green color denotes it is generated by \textcolor{locate}{\textsc{\textbf{Locate}}}, \textcolor{relate}{\textsc{\textbf{Relate}}}, \textcolor{func}{\textsc{\textbf{Func}}}, respectively. }
	\label{fig:example}
\end{figure*}

%% file: tables/sota.tex
\begin{table}[!ht]
\begin{center}
\resizebox{\linewidth}{!}{
\begin{tabular}{|l|c|c|c|c|}
\hline
 & \multicolumn{4}{c|}{MSVD} \\ \hline
Models & B@4 & R & M & C  \\ \hline
MAM-RNN~\cite{li2017mam} &41.3&68.8&32.2&53.9\\
RecNet~\cite{wang2018reconstruction} &52.3&69.8&34.1&80.3\\
MARN~\cite{pei2019memory} &48.4 & 71.9&35.1 & 92.2\\ 
OA-BTG~\cite{zhang2019object} &\textbf{56.9} & -&\underline{36.2}&90.6 \\ 
POS-CG~\cite{wang2019controllable} &52.5 &71.3 &34.1 &92.0 \\
Mixture~\cite{hou2019joint} &52.8 & 71.8&36.1 &87.8 \\ \hline
RMN (S+L) & 52.5 & \underline{72.7} &36.1 & \underline{92.8} \\ 
RMN (H+L) & \underline{54.6} & \textbf{73.4} & \textbf{36.5} & \textbf{94.4} \\
\hline
 \multicolumn{5}{c}{ } \\
\hline
 & \multicolumn{4}{c|}{MSR-VTT} \\ \hline
Models & B@4 & R & M & C \\ \hline
RecNet~\cite{wang2018reconstruction} &39.1&59.3&26.6&42.7\\
MARN~\cite{pei2019memory} &40.4 & 60.7& 28.1& 47.1\\ 
OA-BTG~\cite{zhang2019object} & 41.4& -& 28.2&46.9 \\ 
POS-CG~\cite{wang2019controllable} &42.0 &61.6 &28.2 & 48.7\\
Mixture~\cite{hou2019joint} &\underline{42.3} &\textbf{62.8} & \textbf{29.7}&\underline{49.1} \\ \hline
RMN (S+L)& 42.1 & 61.3 & 28.3 & \underline{49.1} \\ 
RMN (H+L) & \textbf{42.5} & \underline{61.6} & \underline{28.4} & \textbf{49.6} \\ \hline

\end{tabular}
}
\end{center}
\caption{Comparing with the state-of-the-art on MSVD and MSR-VTT datasets. B@4, R, M, C denote BLEU-4, ROUGE\_L, METEOR, CIDEr, respectively. The highest score is highlighted in bold and the second highest is underlined.}
\label{table:sota}
\end{table}

%% file: 5_conclusion.tex
\section{Conclusion}
In this paper, we proposed a novel reasoning neural module networks (RMN) for video captioning that performs visual reasoning on each step along the generation process. Specifically, we designed three sophisticated reasoning modules for spatio-temporal visual reasoning. To dynamically compose the reasoning modules, we proposed a discrete module selector which is trained by a linguistic loss with a Gumbel approximation. Extensive experiments verified the effectiveness of the proposed RMN, and the qualitative results indicated the caption generation process is explicit and explainable.

\section*{Acknowledgments}
This work was supported by the National Key R\&D Program of China under Grant 2017YFB1300201, the National Natural Science Foundation of China (NSFC) under Grants U19B2038, 61620106009 and 61725203 as well as the Fundamental Research Funds for the Central Universities under Grant WK2100100030. 